%% file: root.tex
\documentclass[letter, 10pt, conference]{ieeeconf}      %

\IEEEoverridecommandlockouts                              %

\usepackage{graphicx}
\usepackage{hyperref}
\usepackage{float}
\usepackage{multicol,caption}
\usepackage{cite}
\usepackage{multirow}
\usepackage[square,numbers]{natbib}
\usepackage{bm}
\usepackage{xcolor}
\usepackage{algorithm}
\usepackage[noend]{algorithmic}
\usepackage{amsmath}
\usepackage{amssymb}
\usepackage{booktabs}

\title{\LARGE \bf
Design and Control Co-Optimization for Automated Design Iteration of Dexterous Anthropomorphic Soft Robotic Hands
}

\definecolor{mydarkblue}{rgb}{0,0.08,0.45}
\hypersetup{
    colorlinks=true,
    linkcolor=mydarkblue,
    citecolor=mydarkblue,
    filecolor=mydarkblue,
    urlcolor=mydarkblue,
    }

\author{Pragna Mannam$^{1*}$, Xingyu Liu$^{2*}$, Ding Zhao$^{2}$, Jean Oh$^{1}$, and Nancy Pollard$^{1}$%
\thanks{*Equal Contribution}%
\thanks{$^{1}$Robotics Institute,
Carnegie Mellon University, PA 15213, USA
        {\tt\small pmannam@andrew.cmu.edu}}%
\thanks{$^{2}$Department of Mechanical Engineering,
        Carnegie Mellon University, PA 15213, USA
        {\tt\small xingyul3@andrew.cmu.edu}}%
}

\newcommand{\dash}{\textsc{DASH}}

\newcommand{\vone}{\textsc{v1}}
\newcommand{\vtwo}{\textsc{v2}}
\newcommand{\vthree}{\textsc{v3}}
\newcommand{\vfour}{\textsc{v4}}
\newcommand{\vfive}{\textsc{v5}}
\newcommand{\vsix}{\textsc{v6}}
\newcommand{\vseven}{\textsc{v7}}

\begin{document}

\maketitle
\thispagestyle{empty}
\pagestyle{empty}

\begin{abstract}
We automate soft robotic hand design iteration by co-optimizing design and control policy for dexterous manipulation skills in simulation. Our design iteration pipeline combines genetic algorithms and policy transfer to learn control policies for nearly 400 hand designs, testing grasp quality under external force disturbances. We validate the optimized designs in the real world through teleoperation of pickup and reorient manipulation tasks. Our real world evaluation, from over 900 teleoperated tasks, shows that the trend in design performance in simulation resembles that of the real world. Furthermore, we show that optimized hand designs from our approach outperform existing soft robot hands from prior work in the real world. The results highlight the usefulness of simulation in guiding parameter choices for anthropomorphic soft robotic hand systems, and the effectiveness of our automated design iteration approach, despite the sim-to-real gap.
\end{abstract}

\input{Sections/Intro}

\input{Sections/RelatedWork}

\input{Sections/RobotHand}
\input{Sections/DesignOptimization}

\input{Sections/DesignEvaluation}
\input{Sections/RobotExperiments}

\input{Sections/Discussion}

\addtolength{\textheight}{0cm}   %

\begin{small}
\textbf{Acknowledgment}
This work was supported by the Technology Innovation Program (20018112, Development of autonomous manipulation and gripping technology using imitation learning based on visual-tactile sensing) funded by the Ministry of Trade, Industry \& Energy (MOTIE, Korea).
This material is based upon work supported in part by the AI Research Institutes program supported by NSF and USDA-NIFA under AI Institute: for Resilient Agriculture, Award No. 2021-67021-35329.   
\end{small}

\bibliographystyle{IEEEtran}
\bibliography{IEEEexample}

\end{document}

%% file: Sections/Intro.tex
\section{INTRODUCTION}
\label{sec:Intro}

Humans have evolved to master manipulation of human-designed objects and tools. 
Similarly, robot hands have been designed over many years to progress towards human-like dexterity~\cite{piazza2019century}. 
Multi-fingered robotic hands can vary in degrees-of-freedom, actuation, and more. 
It can be time-consuming to evaluate a proposed hand design along common metrics such as grasp taxonomies or manipulation task success~\cite{feix, YCB, Cruciani_2020, yang2020benchmarking}. 
While some suite of tasks aim to evaluate dexterous manipulation~\cite{dapg, morgan2019learning,cruciani_2018_dexterous}, it is difficult to learn dexterous skills for new hand designs which limits the utilization of dexterous skill benchmarks for evaluation. Thus, we want to efficiently learn dexterous manipulation skills on new hand morphologies in order to evaluate numerous designs rapidly for design iteration.

Testing robot hands on their downstream tasks, such as picking up a cellphone from the table, can be difficult to execute for hundreds of designs since each design requires a different control policy. However, recent works have shown that transferring control policies across evolutions of robotic designs from a source to target robot improves efficiency for learning new policies~\cite{liu2022revolver, liu2022herd}. 
Continuous evolutionary models allow for transferring policies from expert policies on source robots to intermediate robots instead of learning new policies for target robots. While these approaches focus on policy optimization, we show that we can apply this framework to design and policy co-optimization for automating design iteration of robotic hands. 

\begin{figure}[t]
    \centering
    \includegraphics[width=\linewidth]{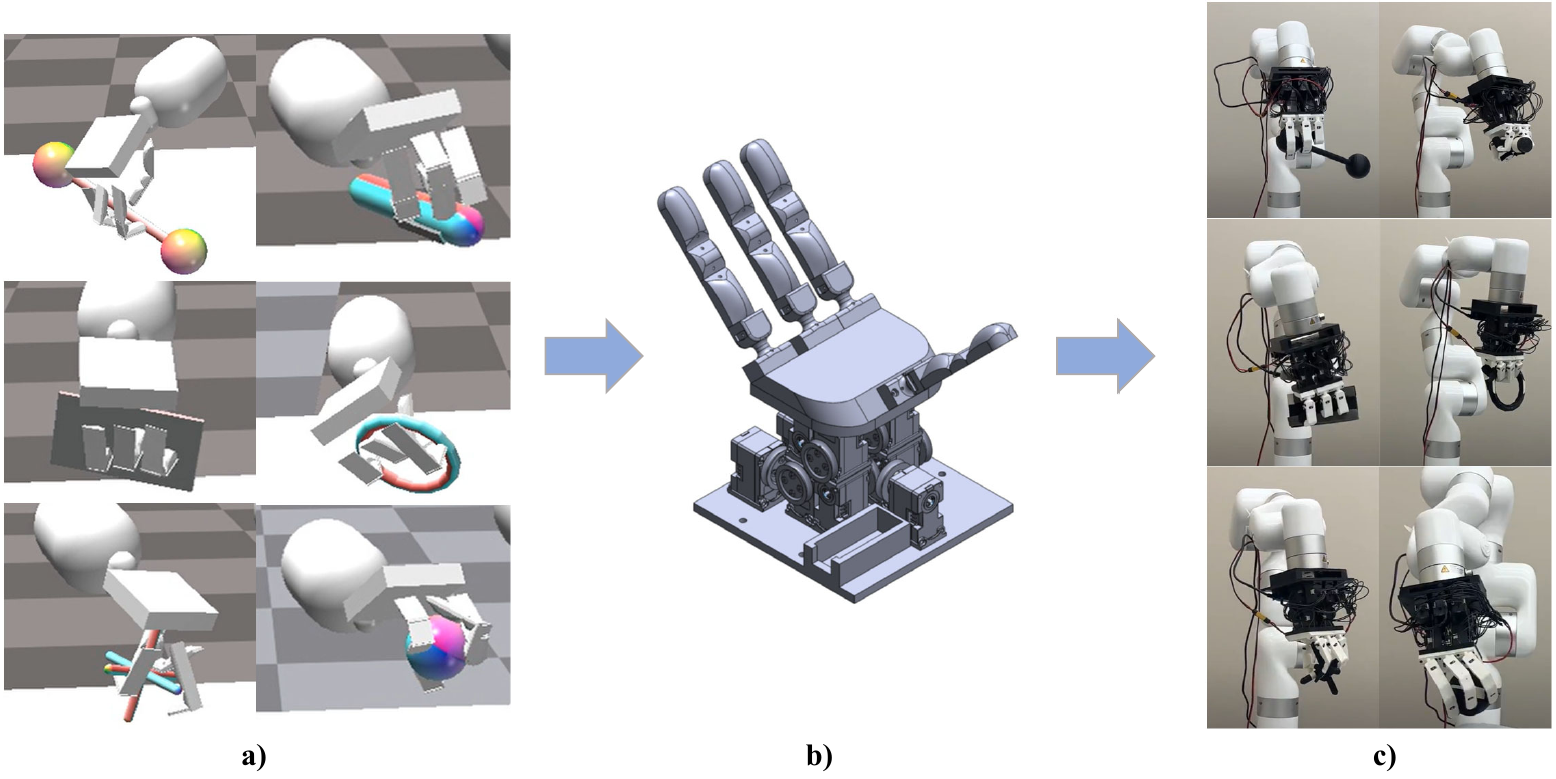}
    \caption{\small Our approach includes: a) hand design optimization in simulation using genetic algorithms and policy transfer; b) CAD for 3D-printing optimized hand that outperforms other hand designs in simulation; and c) real-world design evaluation of optimized hand using teleoperation on the same set of manipulation tasks.}
    \label{fig:teaser}
    \vspace{-0.5cm}
\end{figure}

Using genetic algorithms and robot interpolation, we can explore dexterous hand designs for a set of manipulation tasks. Genetic algorithms have been used for optimization problems including mobile robots and trajectory planning~\cite{davidor1991genetic,manikas2007genetic,vstevo2014optimization}. Robotic interpolation allows for finding the morphology and kinematics of intermediate robots~\cite{liu2022herd,liu2022revolver,liu2023meta}. We utilize both of these concepts to generate new robotic hand designs and learn control policies for them in order to find an optimized top-performing robotic hand design.  

Through learning dexterous manipulation skills on hundreds of robotic hand designs in simulation, we can automate the design iteration of our robotic hands. 
As shown in Figure~\ref{fig:teaser}a, we use genetic algorithms to start from a set of hand design candidates from prior work~\cite{mannam2023framework}, allowing for crossovers and mutations to exploit and explore design features, such as finger arrangement and length, that allow for successful manipulation of six different objects in pick up and reorient manipulation tasks. These dexterous manipulation tasks allow for differentiating hand design performance among $396$ anthropomorphic hand designs that were generated. Subsequently, we fabricate the best-performing hand designs from simulation using 3D-printing (Figure~\ref{fig:teaser}b). 
Using the same six objects in the real world, we use teleoperation to pick up and reorient the objects in-hand, as depicted in Figure~\ref{fig:teaser}c, to evaluate whether the optimized hand designs can succeed at the dexterous manipulation tasks. 

Our hypothesis that the trend in performance of these hand designs in simulation resembles real world performance is supported by our results from more than $900$ real-world teleoperated manipulation experiments. 
In addition, our design and policy co-optimization approach results in two optimized soft robot hand designs, from simulation, that outperform existing soft hand designs, from prior work~\cite{mannam2023framework}, in real world evaluation, despite the considerable sim-to-real gap.
In summary, the contributions of this work include:
\begin{itemize}
    \item An approach for design and policy co-optimization for soft robot hands using genetic algorithm and policy transfer in simulation.
    \item Generation of new hand designs, using the above approach, in simulation that can be fabricated in the real world as tendon-driven soft hands.
    \item Teleoperated evaluation of two optimized fabricated hand designs in the real-world to show that they outperform existing soft hands from prior work~\cite{mannam2023framework}.
\end{itemize}

%% file: Sections/RelatedWork.tex
\section{Related Work}
\label{sec:Related_Work}

Design and control of a robot are intrinsically linked. Hence, jointly co-optimizing design and control policies of robots can be done to avoid determining control policies for every single design candidate. This idea has been explored in a wide range of areas extending from locomotion~\cite{whitman2020modular} to manipulation~\cite{chen2021codesigning}. Recent work have leveraged data-driven and learning-based approaches for co-optimization using concurrent networks~\cite{chen2022c2codesign}. Co-optimization can also be posed as an optimal control problem and has been applied to both manipulation and locomotion using this formulation~\cite{ha2017joint}. Similarly, reinforcement learning can be used to evolve legged robots and their gait towards an optimal design and control policy~\cite{schaff2019jointly, whitman2023learning}. Our approach also uses a reinforcement learning based approach to co-optimize both the design of an anthropomorphic soft robot hand and a control policy for dexterous manipulation tasks.

For design and control co-optimization of robot hands, previous works have used gradient-based approaches and evolutionary methods~\cite{chen2021codesigning}. Xu et. al.~\cite{xu2021endtoend} show that using gradient-based optimization methods outperform both gradient-free methods and model-free reinforcement learning approaches for co-optimization. 
On the other hand, evolutionary methods are a scalable solution to reinforcement learning~\cite{salimans2017evolution}. Continuous robot evolution and human demonstrations can transfer control policies from a five-fingered dexterous robot hand to a two-finger gripper~\cite{liu2022herd}. Furthermore, evolutionary algorithms can be used to co-optimize hand design and control for any manipulation task through optimization of joint limit parameters for increased robustness~\cite{meixner2019automated}. While this approach aims to find the simplest hand design for the task, targeting grasping strategies for specific manipulators,like a soft hand, can adapt existing control strategies or invent new control strategies to co-optimize design and control~\cite{deimel2017automated}. In a similar fashion, our approach uses genetic algorithm for exploring new hand designs and policy transfer to learn new policies for hand designs efficiently in simulation.

Optimizing on robotic hand designs in simulation can pose difficulties for sim-to-real transfer when executing the control policies in the real world. Instead, we use teleoperation setups similar to those previously used for imitation learning data collection~\cite{dime_lerrel, videodex}. Our real world evaluation setup is similar to our previous work~\cite{mannam2023framework}, where we use teleoperation to evaluate the capabilities and limitations of our soft robot hands and their designs. However, our prior work used manual design iteration and did not leverage simulation to optimize hand designs. Our manipulation task setup is similar to the tool positioning task used for chaining multiple dexterous tasks for long-horizon task goals~\cite{chen2023sequential}, where an object is first grasped from the table and then reoriented in-hand to a final desired pose. Contrary to~\cite{chen2023sequential}, we evaluate the same task in the real-world using teleoperation for our generated hand designs. Finally, our approach uses a qualitiative metric, similar to one used in~\cite{kim2013physically}, that captures the quality of the grasp for evaluating the teleoperated manipulation tasks for each of the optimized hand designs in the real world.

%% file: Sections/RobotHand.tex
\section{Dexterous Anthropomorphic Soft Hand}
\label{sec:Robotic_Hand}

\begin{figure}
    \centering
    \includegraphics[width=0.8\linewidth]{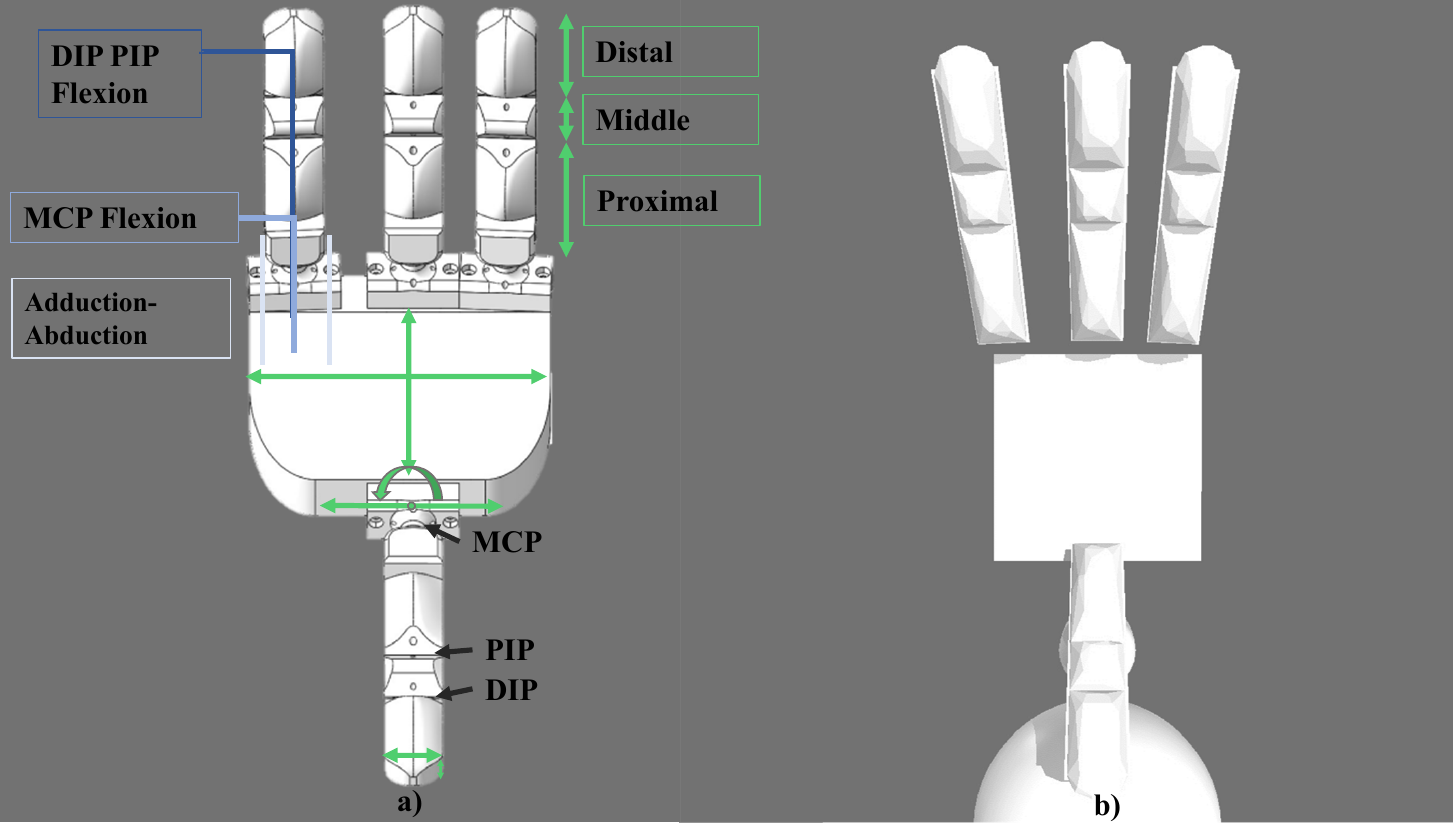}
    \caption{\small a) CAD of top-performing optimized hand  
\vseven{} with DIP, PIP, and MCP joints labelled, as well as tendon placements along the finger shown in blue. The distal, middle, and proximal phalanges are also labelled in green to show part of the hand design parameters. b) Visualization of simulated top-performing hand design \vseven{}. }
    \vspace{-.5cm}
    \label{fig:v7_cad_visual}
\end{figure}

Our robotic hand designs are simulated as rigid robotic hands but fabricated as soft tendon-driven hands using 3D-printing. We implement the robotic hand designs as soft robots for safety during interactions with objects and the environment as well as robustness to error due to compliance~\cite{bonilla2014grasping}. Bauer et al.~\cite{bauer2022towards} show that crease-like deformations in soft robotic hands can be kinematically approximated as rigid joints for design validation. Similarly, we use \dash{}, a Dexterous Anthropomorphic Soft Hand~\cite{mannam2023framework} with joint-like creases that is customizable, low-cost, easy to rapidly prototype and evaluate in the real world using teleoperation.

\subsubsection{DASH Morphology: }
From previous work~\cite{mannam2023framework}, we use five existing design iterations of \dash{}, named \vone{}, \vtwo{}, \vthree{}. \vfour{}, and \vfive{} to start our design optimization process.
Each version of \dash{} incorporates four fingers: the thumb, index, middle, and ring fingers. To ensure modularity, every finger, including the thumb, shares an identical design. Each of these fingers consists of three joints along its length, moving from the base towards the fingertip: the metacarpophalangeal (MCP) joint, proximal interphalangeal (PIP) joint, and distal interphalangeal (DIP) joint. The joints for each individual finger are illustrated in Figure~\ref{fig:v7_cad_visual}.

Each finger has four tendons (as shown in Figure~\ref{fig:v7_cad_visual}), two are situated along the sides of the MCP joint, near the palm, responsible for abduction and adduction—permitting the fingers to move both closer together and farther apart. These two tendons are attached to a bi-directional pulley actuated by a single motor, collectively referred to as the adduction-abduction tendon.
The MCP flexion tendon is used for the forward folding of the finger at the MCP joint, perpendicular to the abduction-adduction tendon's motion. Lastly, the DIP PIP flexion tendon extends through the entire length of the finger in order to fully curl the finger inwards.

\subsubsection{Fabrication and Customization: }
In Figure~\ref{fig:teaser}(c), the fingers and palm were 3D-printed from Ninjaflex Edge (83A shore hardness)~\cite{ninjaflex} using a FlashForge Creator Pro 2 printer. Tendon routing and finger morphology can be easily changed to 3D-print a new robotic hand design, facilitating rapid prototyping and evaluating. We use $12$ Dynamixel XC330-M288-T motors~\cite{dynamixel} in the motor housing which is 3D-printed from PLA and shown below the palm in Figure~\ref{fig:teaser}(b). 
Our real robot experiments used the robot hand mounted to an xArm7 robotic arm~\cite{xarm7}. 

\subsubsection{Design Parameters: }
While 3D-printing our soft dexterous hands allows us endless possibilities in terms of design, we constrain the design parameters to ones that allow for significant manipulation behavior changes. In prior work~\cite{mannam2023framework}, the design changes among \dash{} \vone{} to \vfive{} included parameters such as palm size, finger length, joint height and width, thumb arrangement, and fingertip shape. 

For our simulated design optimization, we restrict the design changes to the following parameters illustrated in Figure~\ref{fig:v7_cad_visual}a: palm size, finger links lengths (proximal, middle, distal phalanges' lengths), finger width, finger thickness, finger position, and finger orientation. The ranges of these parameter values are detailed in Table~\ref{tab:hand_parameters}.
Figure~\ref{fig:v7_cad_visual}b shows the visualization of the same hand CAD shown in Figure~\ref{fig:v7_cad_visual}a in simulation where the hand parameters match for both hands. 
Tendon arrangements and actuation were integrated into the design after determining the optimized hand design. %

\begin{table}[t]
\centering
\scriptsize
\setlength{\tabcolsep}{3pt}
\begin{tabular}{ *2l *6c }    \toprule
\multicolumn{2}{l}{\emph{{\textbf{Design Params}}}} & \emph{Min} & \emph{Max} & \emph{\vthree{}} & \emph{\vfive{}} & \emph{\vsix{}*} &  \emph{\vseven{}*} \\ \midrule
\multicolumn{2}{l}{Palm width} & 69 & 99 & 84 & 84 & 92 & 92 \\
\midrule
\multicolumn{2}{l}{Palm height} & 69 & 99 & 84 & 84 & 74 & 74\\
\midrule
\multirow{3}{*}{Position} & ff & (8, 64) & (48, 84)  & (28, 84) & (28, 84)  & (28, 84) & (29, 83) \\
& mf & (-20, 64) & (20, 84) & (0, 84) & (0, 84) & (0, 84) & (0, 84)\\
& rf & (-56, 64) & (-16, 84) & (-28, 84) & (-28, 84) & (-36, 84) & (-36, 83) \\
\midrule
\multirow{3}{*}{Orientation} & ff & 0 & 45 & 0 & 0 & 0 & 2.9 \\
& mf & -35 & 35 & 0 & 0 & 0 & 0\\
& rf & -45 & 0 & 0 & 0 & 0 & -2.9 \\
\midrule
\multicolumn{2}{l}{Proximal length}  & 35 & 55  & 45 & 45 & 45 & 45 \\
\midrule
\multicolumn{2}{l}{Middle length}  & 8 & 28  & 20 & 20 & 18 & 18\\
\midrule
\multicolumn{2}{l}{Distal length}   & 25 & 45 & 35 & 35 & 35 & 35 \\
\bottomrule
 \hline
\end{tabular}
\caption{\small \label{tab:hand_parameters} Hand design parameter ranges tested in sim to find the optimized designs where ff, mf, and rf refer to index, middle, and ring finger, respectively.
Unit is mm for lengths and deg for angles. Designs with * are sim optimized designs.}
\vspace{-.5cm}
\end{table}

%% file: Sections/DesignOptimization.tex
\section{Design and Policy Co-Optimization}
\label{sec:Design_Optimization}

We assume the hand robot can be defined by its design parameters.
This assumption is true for the design space of our DASH hand where we fix the topology of kinematic connections of bodies and joints.
In this section, we describe our proposed method for automatic optimization of the DASH hand design parameters. 

\subsection{Problem Definition and Preliminaries}

We formulate the robotic manipulation task as a Markov Decision Process (MDP) specified by a tuple $(\mathcal{S}, \mathcal{A}, \mathcal{T}, R, \gamma)$, where 
$\mathcal{S} \subseteq \mathbb{R}^S$ is the state space, $\mathcal{A} \subseteq \mathbb{R}^A$ is the action space, $\mathcal{T}: \mathcal{S} \times \mathcal{A} \rightarrow \mathcal{S} $ is the transition function, $R: \mathcal{S} \times \mathcal{A} \rightarrow \mathbb{R}$ is the reward function, and $\gamma \in [0,1]$ is the discount factor.
A policy $\pi: \mathcal{S} \rightarrow \mathcal{A}$ maps a state to an action where $\pi(a|s)$ is the probability of choosing action $a$ at state $s$.
We assume that all possible robot designs share the same state space $\mathcal{S}$ and action space $\mathcal{A}$.
We assume a robot can be defined by $D$ independent design parameters that only impact the transition dynamics $\mathcal{T}$.
Suppose $\rho^{\pi,\theta} = \sum_{t=0}^\infty \gamma^t R(s_t,a_t)$ is the episode discounted reward when a robot with design parameters $\theta \in \mathbb{R}^D$ executes policy $\pi$.
The optimal policy $\pi^*_\theta = \mathop{\arg\max}_\pi \mathbb{E}[\rho^{\pi, \theta}]$ on robot $\theta$ is the one that maximizes the expected value of $\rho^{\pi,\theta}$.

Given the MDP for a certain manipulation task, the goal of our problem is to find the optimal design parameters of the robot design defined as
\vspace{-0.1cm}
\begin{equation}
    \theta^* = \mathop{\arg\max}_{\theta} \rho^{\pi_\theta^*, \theta}
\end{equation}
This means robot designs should be compared by the performance of their well-trained expert policy.
The difficulty of this optimization problem stems from two aspects. 
First, finding the expert policy $\pi_\theta^*$ by training policies from scratch for each robot design $\theta$ is computationally expensive.
Second, the high-dimensional robot design space has an exponential complexity with respect to the dimension $D$.

To address these challenges in robot design optimization, we aim to improve policy optimization efficiency while allowing global search of robot design parameters.
Recent advances in interpolation-based policy transfer \cite{liu2022revolver,liu2022herd} inspire us to address this problem from the perspective of policy transfer.
The intuition is that, if two robots are similar in their hardware configuration, their optimal policy should also be similar. 
Therefore, given a robot $\theta_\text{source} \in \mathbb{R}^D$ with an expert policy $\pi_{\theta_\text{source}}^*$ and another significantly different robot $\theta_\text{target} \in \mathbb{R}^D$, it may be possible to interpolate the two robots by producing a sequence of intermediate robots $\theta_1, \theta_2, \ldots, \theta_{K-1}, \theta_K$ that continuously changes from $\theta_1 = \theta_\text{source}$ to $\theta_K = \theta_\text{target}$.
These intermediate robots act as the stepping stones for transferring the policy.
It takes $K$ training phases to transfer the policy where in $i$-th phase, the policy is fine-tuned on $\theta_i$ until convergence.
Since the difference between any two consecutive robots $\theta_i$ and $\theta_{i+1}$ in the sequence is sufficiently small, the overall training overhead for transferring of policy is much smaller than training the policy on $\theta_\text{target}$ from scratch.

\subsection{Genetic Algorithm and Policy Transfer for Design Optimization}

Our method leverages  the above idea to efficiently transfer the expert policies on existing robot designs to multiple robot design candidates that are iteratively searched and trained.

Specifically, given the pool of $C$ existing robot designs $ \mathcal{D}_e = \{ \theta_i \in 
\mathbb{R}^D \mid i = 1, 2, \ldots, C \}$ and their respective well-trained expert policy $\pi_{\theta_i}^*$, we randomly generate a new robot  design candidate $\theta_\text{new} \in 
\mathbb{R}^D$. 
To obtain the optimal policy on the new design candidate, it is natural to find the closest existing robot design with expert policy as the starting point. %
Suppose $\theta_s = \mathop{\arg\min}_{\theta \in \mathcal{D}_e} || \theta - \theta_{\text{new}} ||$ is the robot with the smallest hardware difference to $\theta_\text{new}$ in $\mathcal{D}_e$.
Then the next robot stepping stones can be obtained by moving the design parameter $\theta_s$ towards $\theta_\text{new}$.
We employ a small step size $\xi \in \mathbb{R}^+$ for the change of design parameter.

How can we randomly generate new hand design candidates?
Given large $D$, it is computationally intractable to iterate all possible design candidates.
Our intuition is that  good robot designs are usually superior due to certain ``traits'', e.g. long fingers, wide motion ranges.
It makes sense to share the good traits with other designs to search potentially better designs.
Genetic algorithm is an optimization scheme that is capable of achieving this.
It contains two parts: crossover and mutation.
In every design exploration phase, we randomly sample two existing robot design candidates $\theta_1, \theta_2$ from $\mathcal{D}_e$.
We perform element-wise random crossover on $\theta_1$ and $\theta_2$ to generate a new $\theta_\text{new}$.
Then we mutate $\theta_\text{new}$ by adding a noise term that is uniformly sampled from $[-\theta_M, \theta_M]$, where $\theta_M \in {\mathbb{R}^+}^D$.
In this way, $\theta_\text{new}$ can inherit the good traits from existing good designs and also explore in the design space for potential improvement over the current state of the art.

Note that the robot design pool $\mathcal{D}_e$ is not static.
To fully re-use the previously searched good designs, $\mathcal{D}_e$ is dynamically populated with newly found designs, including the designs acting as policy transfer stepping stones.
However, in order for the genetic algorithm to find better designs, $\mathcal{D}_e$ should only contain the most elite population of robot designs.
Therefore, we set a threshold of robot policy performance $q$ for deciding whether a robot design is good enough to be added to the pool $\mathcal{D}_e$.
The threshold $q$ is selected heuristically.
Our overall design and policy co-optimization method is illustrated in Algorithm \ref{alg:design:opt}.

\subsection{Remarks}
Our robot design optimization scheme does not guarantee convergence as genetic algorithms are not guaranteed to converge to a global optima.
However, with sufficient amount of computation, our method can cover enough good design space to result in optimized designs, which can serve as good candidates for real-world fabrication of our DASH hand.

Note that it is possible that in some iterations, the randomly generated new design candidate $\theta_\text{new}$ is inferior with low rewards.
In this case, during policy transfer, upon reaching a certain intermediate robot $\theta$ along the path, the well-trained expert policy $\pi_\theta^*$ will no longer reach the performance threshold $q$.
Though this bad design does not provide new members for $\mathcal{D}_e$, it can still provide valuable information on what design parameter changes might be useful.

\input{algo}

%% file: algo.tex
\begin{algorithm}[t]
\small
\caption{\small{Design and Policy Co-Optimization}}
\label{alg:design:opt}
\begin{algorithmic}[1]
\STATE{\textbf{Notation Summary:}
$\theta_i \in 
\mathbb{R}^D$: design parameter of $i$-th robot;
$\pi_{\theta_i}^*$: expert policy on $i$-th robot; 
$\xi \in \mathbb{R}^+$: evolution step size;
$q \in \mathbb{R}$: reward threshold;
$\mathcal{D}_e$: the set of robots with expert policy;
$\theta_M \in \mathbb{R}^D$: element-wise mutation range.
$\theta_U, \theta_L \in \mathbb{R}^D$: element-wise upper and lower bounds of design parameter.
}
\\
\hrulefill
\\
\STATE{ $\mathcal{D}_e \leftarrow \{ \theta_1, \theta_2, \ldots, \theta_C \}$; }
\FOR{$i$ \textbf{in} $1, 2, \ldots, N$}
    \STATE{\textcolor{darkgray}{// sample new robot design candidate}}
    \STATE{Sample $\theta_1, \theta_2 \sim \mathcal{D}_e$, mutation noise $n \sim U([-\theta_M, \theta_M])$; }
    \STATE{$\theta_{\text{new}} \leftarrow \text{\texttt{random\_crossover}}(\theta_1, \theta_2) + n$; }
    \STATE{$\theta_{\text{new}} \leftarrow \text{MAX}(\text{MIN}(\theta_{\text{new}}, \theta_U), \theta_L)$; \textcolor{darkgray}{// stay in bound} }
    \STATE{\textcolor{darkgray}{// find closest source robot to transfer policy from}}
    \STATE{$\theta_s \leftarrow \mathop{\arg\min}_{\theta \in \mathcal{D}_e} || \theta - \theta_{\text{new}} || $; }
    \STATE{$\theta \leftarrow \theta_s$, $\pi^* \leftarrow \pi_{\theta_s}^*$; }
    \STATE{\textcolor{darkgray}{// transfer the policy by robot interpolation}}
    \WHILE{$||\theta - \theta_{\text{new}}|| > \varepsilon$}
    \STATE{$\theta \leftarrow \theta + \xi \cdot (\theta_{\text{new}} - \theta) / ||\theta_{\text{new}} - \theta|| $; }    
    \STATE{train expert policy $\pi_{\theta}^* \leftarrow \mathop{\arg\max}_{\pi} \mathbb{E} [\rho^{\pi, \theta}] $ by initializing policy with $\pi^*$; }
    \STATE{ $\pi^* \leftarrow \pi_{\theta}^*$}
    \IF{$\mathbb{E} [\rho^{\pi_\theta^*, \theta}] > q$}
        \STATE{$\mathcal{D}_e \leftarrow \mathcal{D}_e \cup \{ \theta \}$; \textcolor{darkgray}{// only keep elite robot candidates} }
    \ENDIF    
    \ENDWHILE
\ENDFOR
\STATE{\textbf{return} $\{ (\theta, \pi_{\theta}^* ) \mid \theta \in \mathcal{D}_e\}$; }
\end{algorithmic}
\end{algorithm}

%% file: Sections/DesignEvaluation.tex
\section{Simulation Benchmark and Evaluation}
\label{sec:Simulation_Evaluation}

\subsection{Benchmark Definition}

Using simulation to optimize the hand design requires designing a benchmark suite whose manipulation task and evaluation metrics can  distinguish between similar hands.
In this section, we propose a benchmark task suite for evaluating dexterous hands.
In contrast to related works that separate object relocation and in-hand reorientation~\cite{dapg}, we combine object relocation and reorientation into one task where the goal is to pick up the object and then reorient the object in-hand to desired relative 6D pose between hand and object.
The success condition is defined as 6D pose being sufficiently close to the goal pose in both rotation and translation components.
In this way, both object grasp ability and hand dexterity can be fully evaluated in a single task.

To evaluate the full potential of the hand designs, we include six objects with diverse geometry in our benchmark suite.
The six objects include \texttt{barbell}, \texttt{board}, \texttt{cross3d}, \texttt{pen}, \texttt{ring} and \texttt{sphere} as illustrated in Figure \ref{fig:teaser}.
The objects are chosen such that the their geometry is diverse enough to evaluate different taxonomy of hand grasping and reorientation.
For example, \texttt{pen} can help evaluate a precision grasp while \texttt{sphere} focuses more on power grasps. 
Each object is instantiated with three different uniform scales: 0.75$\times$, 1$\times$, and 1.25$\times$, resulting in 18 object instances in total.
The hand designs are evaluated on manipulation tasks on all 18 object instances.

\subsection{Evaluation Metrics}

Defining an episode success condition and using the success rate as the evaluation metric has been a popular approach in previous works.
However, using a discrete signal of episode success or failure cannot sufficiently distinguish subtle difference in hand designs.
On the other hand, defining dense and continuous evaluation metrics cannot easily generalize to other tasks and is not straightforward to define for our object relocation and reorientation task.

We propose a new evaluation metric to address the above problem.
During evaluation, we apply an unknown external force $F$ to the object in simulation.
The direction of the external force is random and stays the same throughout the episode.
We then measure the average success rates under different external force magnitudes.
Formally, let $S(\pi, \theta, F)$ be the average success rate of executing policy $\pi$ on robot $\theta$ with unknown external force of magnitude $F$.
Our evaluation metric of robot $\theta$ is defined as
\begin{equation}\label{eq:evaluation}
    M(\theta) = \frac{1}{F_\text{max}}\int_0^{F_\text{max}} S(\pi_\theta^*, \theta, F) dF
\end{equation}
where $F_\text{max}=1$N.
$M(\theta)$ is essentially the area-under-curve (AUC) of success rate $-$ external force curve.
In our experiments, we found that $M(\theta)$ is a more distinguishable metric than naively using success rate $S(\pi_\theta^*, \theta, 0)$.
In practice, the integral in Equation \eqref{eq:evaluation} is approximated by discretized summation. 

\subsection{Other Implementation Details}

To evaluate a large number of hand designs on a variety of tasks, it is imperative to perform large-scale simulation. 
We use Issac Gym \cite{makoviychuk2021isaac}, a GPU-empowered parallel simulator as the physics simulation engine.
For each object instance, we launch 256 simulation environments, resulting in 4,608 environments running in parallel.

\begin{figure}[t]
    \centering
    \includegraphics[width=\linewidth]{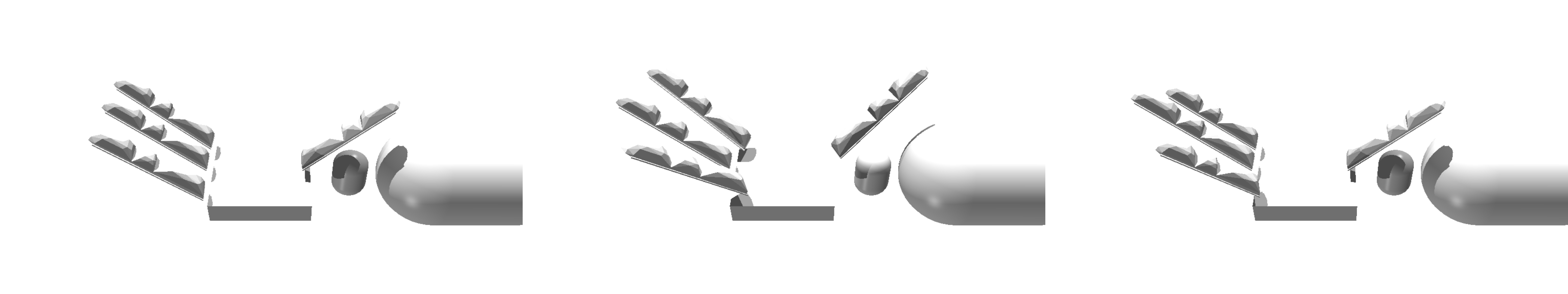}
    \caption{\small Top three optimized hands. From left to right, the success rate AUC (\%) of the hands averaged over all 18 object instances in simulation experiments are 53.32, 53.07 and 52.47 respectively. }
    \label{fig:top:hands}
\vspace{-0.5cm}
\end{figure}

Given 18 different object instances, a naive solution is to train 18 independent policies for each object instance.
This solution does not fully exploit the potential of parallelism of GPU.
Instead, we train a single control policy to manipulate all 18 object instances where the policy is conditioned on the one-hot object ID vectors.
We illustrate three of the top searched hands in Figure \ref{fig:top:hands}.

%% file: Sections/RobotExperiments.tex
\section{Real-World Design Evaluation}
\label{sec:Teleoperation_Evaluation}

 A subset of the simulated hand designs were tested in the real-world using teleoperation. 
 All six objects, \texttt{barbell}, \texttt{board}, \texttt{cross3d}, \texttt{pen}, \texttt{ring} and \texttt{sphere}, were tested in the real-world at 1$\times$ scale. 
 While $256$ randomized goal poses were tested in simulation, we limit our real-world robotic hand design evaluation to 6 goal poses from randomized initial positions for the pick up and reorient manipulation task. We fabricated two optimized hands from simulation from the top 25 hands, where the first ranked hand is the leftmost design shown in Figure~\ref{fig:top:hands}. The tested hands include \dash{} \vthree{}, \vfive{}, and two top 25 ranking optimized hand designs from simulation. We refer to the simulation optimized hands we test as \vsix{} and \vseven{}, where \vseven{} is the best ranked hand in simulation. 

 The hand design parameter values of \vthree{}, \vfive{}, \vsix{}, and \vseven{} are listed in Table~\ref{tab:hand_parameters}. \dash{} \vthree{} and \vfive{} are from previous work using manual design iteration~\cite{mannam2023framework} where \vthree{} has the thumb arranged directly opposing the index finger and \vfive{} has the thumb rotated by $22.5^\circ$ towards the ring finger. Both of these hands have the thumb placed in the bottom right corner of the palm, much like the human hand. The optimized hands, \vsix{} and \vseven{}, have thumbs in the bottom middle of the palm, as shown in Figure~\ref{fig:v7_cad_visual}. These hand designs have a wider but shorter palm with the ring finger offset farther away from the middle finger than the index finger. Unlike \vsix{}, \vseven{} also has the index and ring finger turned toward the middle finger by $3^\circ$. While our optimization did not include fingertip shape, \vthree{} and \vfive{} had a wedge-like and flat fingerpad-like fingertip, respectively. Meanwhile, \vsix{} and \vseven{} had rounded spherical fingertip shapes (as shown in Figure~\ref{fig:teaser}b).

\subsubsection{Experiment Setup}
\label{sec:teleop_setup}

Previous work~\cite{mannam2023framework} has shown that teleoperation removes the necessity to learn control policies for new soft hand designs for quick evaluations of the hand's capabilities in real-world experiments. We use Manus~\cite{manus} VR gloves to teleoperate a 7-DOF robotic arm and our soft hand to evaluate the hand designs in the real-world for various manipulation tasks. Each of the hand designs is calibrated individually to map joint-to-joint control of \dash{} as well as scaling motions such as adduction-abduction motions to be exaggerated from human hand motion for easier teleoperation. Further calibration and teleoperation implementation details can be found in previous work~\cite{mannam2023framework}.

Our real-world teleoperation evaluation setup involves two people. First, we use an experienced teleoperator to perform manipulation tasks with each tested hand for the 6 objects and repeat each goal pose three times. Once a goal pose is achieved, the second person will interactively score the goal pose on a scale of 0 to 1 based on grasp stability. Scores are decided by both people and discussed upon disagreement. These qualitative metrics are similar to related work that interactively tested grasp quality~\cite{kim2013physically}, where: \texttt{1} means a stable grasp (unmovable by small disturbance force), \texttt{0.75} means the object moves by disturbance force but will not drop, \texttt{0.5} means the object can be moved and dropped by disturbance force, \texttt{0.25} means the grasp is fragile and will not be able to carry object, and \texttt{0} means grasping or achieving the goal pose failed.

\begin{figure}
    \centering
    \includegraphics[width=\linewidth]{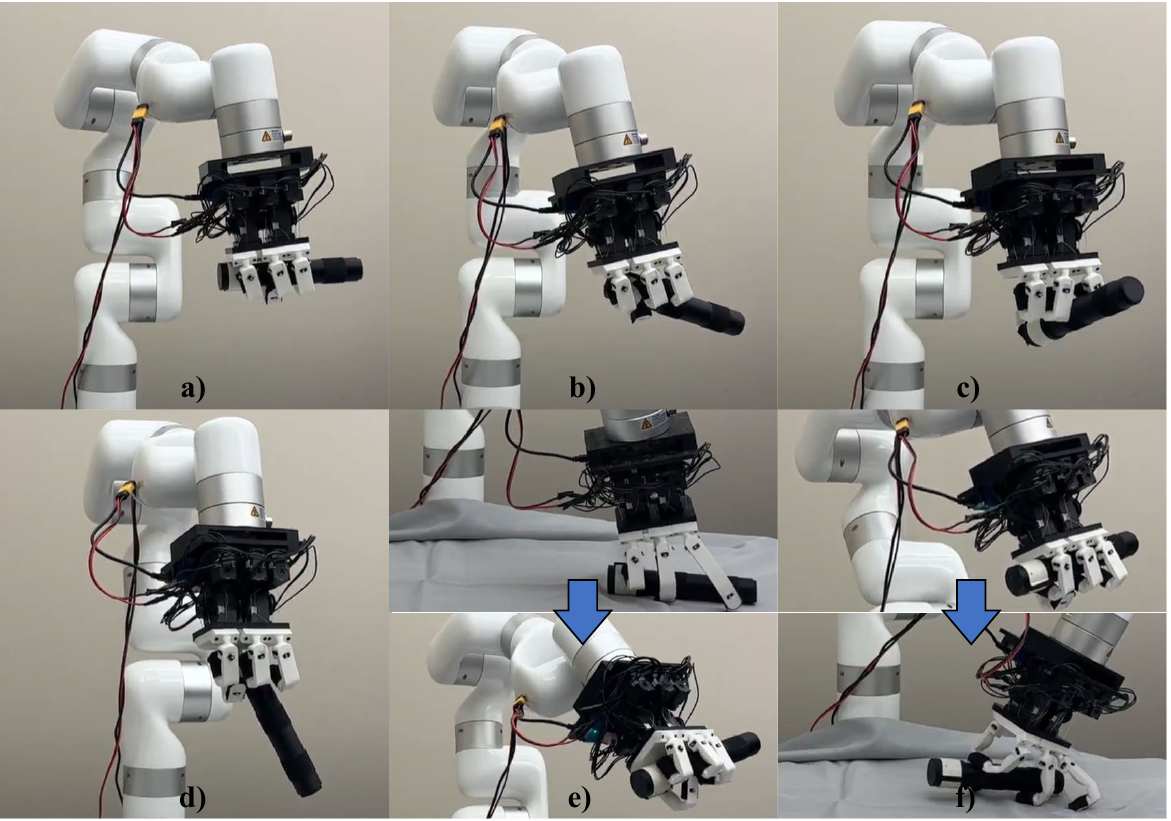}
    \caption{\small The 6 goal poses (shown for \texttt{pen} object) used for real world teleoperated manipulation tasks.}
    \label{fig:goal_poses}
    \vspace{-0.5cm}
\end{figure}

The teleoperator tested \vthree{}, \vfive{}, \vsix{}, and \vseven{} sequentially in one order and repeated each hand in the reverse order to remove the effects of learning through experience teleoperating \dash{} for these tasks. Each hand was tested twice on each object and scores are averaged. The average and maximum scores across the two repetitions of each hand are reported in Tables~\ref{tab:object_results}-~\ref{tab:pose_results}. The objects are augmented with grip tape and foam tape in order to increase friction and gripability of the smooth 3D-printed objects. White tape is used to mark the orientation of the object due to symmetry, as shown in Figure~\ref{fig:goal_poses}. For each object, the six goal poses are referred to as horizontal (\ref{fig:goal_poses}a), tilt $45\deg$ (\ref{fig:goal_poses}b), cigarette grasp (\ref{fig:goal_poses}c), vertical (\ref{fig:goal_poses}d), table to precision grasp to power grasp (\ref{fig:goal_poses}e), and power grasp to precision grasp to table (\ref{fig:goal_poses}f).

These goal poses are shown for the object \texttt{pen} in Figure~\ref{fig:goal_poses}. The last goal pose starts with the object in-hand in a power grasp and requires the object to be gently placed on the table. The quality metrics for goal pose 6 uses a different scoring metric, where \texttt{1} means graceful placement of the object on table within 3cm drop height, \texttt{0.5} means the object initially contacts the table and falls greater than 3cm height, and \texttt{0} means the object falls from the hand greater than 3cm height with no initial contact with the table.

\subsubsection{Teleoperation Evaluation Results}

As shown in Table~\ref{tab:object_results}, the average and maximum scores for each object show that the optimized hands \vsix{} and \vseven{} performed the best. From Table~\ref{tab:pose_results}, we see that \vseven{} struggled with pose 6 the most compared to other goal poses and hand designs. With two fingers rotated inwards towards the middle finger, we observe that \vseven{} can cause unexpected rotations of the object during manipulation. This made \vseven{} easier for power grasps rather than precision grasps. \texttt{cross3d} is our smallest tested object and required precision, which \vfive{}  performed the best on.

\vfive{} has a larger palm and longer fingers than the optimized hands, which performed best on the \texttt{barbell} and \texttt{board} objects as well as outperforming \vsix{} on the tilt goal pose 2. Since goal pose 2 requires the object to be at a 45 degree angle, having the thumb on the bottom right corner of the palm was helpful for \vfive{}. Both \vsix{} and \vseven{} performed similarly on \texttt{barbell}. However,  \vsix{} performed better on \texttt{board}, \texttt{pen}, \texttt{ring} and \texttt{sphere} while \vseven{} performed better on \texttt{cross3d}. Additionally, \vfive{}, \vsix{}, and \vseven{} excelled similarly at the power grasp from the table (\texttt{pose 5}).

Across objects and goal poses, \vthree{} scored lower than other hands. We observed that the combined effect of not being able to fully curl the fingers and its wedge-like fingertips caused \vthree{} to have more unstable grasps compared to the other hands. In previous work~\cite{mannam2023framework}, finger curling ability was improved for \vfive{}, so we see that the scores for \vfive{} are better than \vthree{} for each goal pose in Table~\ref{tab:pose_results}. The easiest goal pose for all hands was the horizontal goal pose 1 and the most difficult pose was starting from power grasp and placing the object on the table (goal pose 6).

\begin{table}[t]
\centering
\begin{tabular}{@{}rrr|rr|rr|rr@{}} \toprule
\emph{\textbf{Objects}} & \multicolumn{2}{c|}{\emph{\vthree{}}} & \multicolumn{2}{c|}{\emph{\vfive{}}} & \multicolumn{2}{c|}{\emph{\vsix{}}} & \multicolumn{2}{c}{\emph{\vseven{}}}  \\
\cmidrule{2-3} \cmidrule{4-5} \cmidrule{6-7} \cmidrule{8-9} 
 & \texttt{avg} & \texttt{max}  & \texttt{avg} & \texttt{max}  & \texttt{avg} & \texttt{max}  & \texttt{avg} & \texttt{max} \\
\midrule
\texttt{barbell} & 0.81 & 0.86 & 0.92 & \textbf{0.99} & \textbf{0.95} & 0.96 & \textbf{0.95} & \textbf{0.99} \\
\texttt{board} & 0.78 & 0.83 & 0.94 & 0.99 & \textbf{0.97} & \textbf{1.00} & 0.92 & 0.99 \\
\texttt{pen}  & 0.78 & 0.79 & 0.79 & 0.89 & \textbf{0.92} & \textbf{0.96} & 0.91 & 0.94 \\
\texttt{ring} & 0.85 & 0.94 & 0.83 & 0.85 & \textbf{0.92} & \textbf{0.96} & 0.90 & 0.90 \\
\texttt{cross3d} & 0.87 & 0.90 & 0.88 & \textbf{0.99} & 0.88 & 0.93 & \textbf{0.94} & 0.94 \\
\texttt{sphere} & 0.67 & 0.78 & 0.83 & 0.92 & \textbf{0.96} & \textbf{1.00} & 0.83 & 0.85 \\
\midrule
\texttt{avg} & 0.79	& 0.85	& 0.87&	0.94 &	\textbf{0.93}&	\textbf{0.97}&	0.91&	0.94 \\
\bottomrule
\end{tabular}
\caption{\small \label{tab:object_results} Teleoperation evaluation results for pick up and reorient tasks for each of the six objects on all goal poses using different hand designs. The above table shows both the average and max grasp quality score achieved.}
\end{table}

\begin{table}[t]
\centering
\begin{tabular}{@{}rrr|rr|rr|rr@{}} \toprule
\emph{\textbf{Goal Pose}} & \multicolumn{2}{c|}{\emph{\vthree{}}} & \multicolumn{2}{c|}{\emph{\vfive{}}} & \multicolumn{2}{c|}{\emph{\vsix{}}} & \multicolumn{2}{c}{\emph{\vseven{}}}  \\
\cmidrule{2-3} \cmidrule{4-5} \cmidrule{6-7} \cmidrule{8-9} 
 & \texttt{avg} & \texttt{max}  & \texttt{avg} & \texttt{max}  & \texttt{avg} & \texttt{max}  & \texttt{avg} & \texttt{max} \\
\midrule
\texttt{pose 1} & 0.86 & 0.93 & 0.94 & \textbf{1.00} & \textbf{0.98} & 0.99 & 0.97 & 0.99 \\
\texttt{pose 2} & 0.80 & 0.88 & \textbf{0.92 }& \textbf{0.99} & 0.91 & 0.92 & 0.88 & 0.94 \\
\texttt{pose 3} & 0.85 & 0.86 & 0.80 & 0.89 & 0.89 & \textbf{0.99} & \textbf{0.94} & 0.96 \\
\texttt{pose 4} & 0.80 & 0.86 & 0.79 & 0.81 & 0.91 & \textbf{0.96} & \textbf{0.93} & 0.94 \\
\texttt{pose 5} & 0.73 & 0.83 & 0.92 & 0.99 & 0.95 & \textbf{1.00} & \textbf{0.97} & 0.99 \\
\texttt{pose 6} & 0.72 & 0.78 & 0.81 & 0.94 & \textbf{0.96} & \textbf{1.00} & 0.76 & 0.78 \\
\midrule
\texttt{avg} & 0.79 &	0.86 &	0.86 &	0.94 &	\textbf{0.93} &	\textbf{0.98} &	0.91 &	0.93 \\
\bottomrule
\end{tabular}
\caption{\small \label{tab:pose_results} Teleoperation evaluation results for pick up and reorient tasks for each of the goal poses on all six objects using different hand designs. The above table shows both the average and max grasp quality score achieved.}
\vspace{-.5cm}
\end{table}

\subsubsection{Simulation vs. Real Results}
\begin{figure}
    \centering
    \small
    \includegraphics[width=\linewidth]{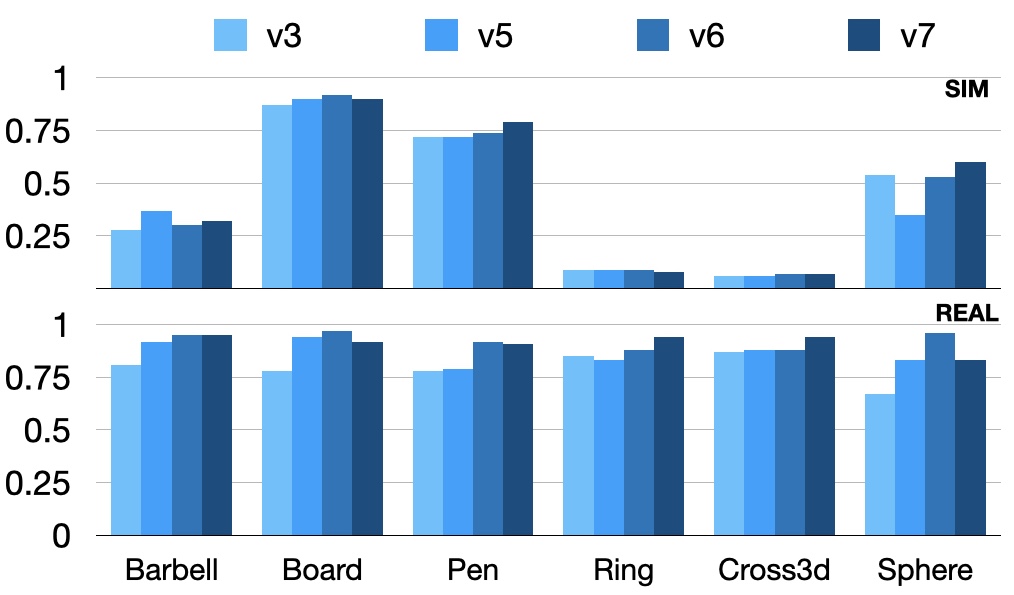}
    \caption{\small (Top) simulation evaluation results averaged for pick up and reorient tasks for each of the six objects at scale $1\times$ on randomized goal poses with AUC success rate given by Equation~\eqref{eq:evaluation}; and (Bottom) Real-world teleoperation evaluation averaged on same objects for six goal poses using grasp quality metric explained in Section~\ref{sec:teleop_setup}.}
    \label{fig:sim_real_graph}
    \vspace{-0.5cm}
\end{figure}

If we look at the performance by object, simulation and teleoperation results follow a similar trend. 
As shown in Figure~\ref{fig:sim_real_graph}, both optimized hand designs \vsix{} and \vseven{} outperform manually iterated designs \vthree{} and \vfive{} on most of the objects in simulation and real world evaluation. 
For simulation and real world setups, \vseven{} performs best on \texttt{barbell} and \texttt{cross3d} objects and \vsix{} performs best on \texttt{board}. 
However, \vsix{} has the highest average score across all objects for real world experiments and \vseven{} has the highest average score across all objects for simulation experiments. 
In simulation, \vseven{} performs best on \texttt{barbell}, \texttt{pen}, and \texttt{sphere}. 
In teleoperation evaluation, \vsix{} performs best on \texttt{board}, \texttt{pen}, and \texttt{sphere}.
And \vsix{} and \vseven{} score similarly for \texttt{barbell} and \texttt{cross3d} in real world and simulation setups, respectively. 
While real-world experiments showed comparable scores on all six objects, simulation results have a high variation depending on objects. 
\texttt{ring} and \texttt{cross3d} have the lowest scores and \texttt{board} and \texttt{pen} have the highest scores. 
In teleoperation experiments, \texttt{pen} and \texttt{sphere} have the lowest scores and \texttt{barbell} and \texttt{board} have the highest scores, but by a small margin. 
This correlates with our observation that the compliance of our soft hands helped most with \texttt{ring} and \texttt{cross3d} objects for conforming to the object for stable grasps. 
Overall, the performance of hand designs simulated as rigid bodies correlate with the performance of our teleoperation evaluation results with \vsix{} and \vseven{} performing best on most test objects. 

\subsubsection{Evaluation of Task Generalization}
To further understand the capabilities and limitations of our best ranking hand design \vseven{}, we evaluate \vseven{} on \dash{}-30, a suite of 30 manipulation tasks from prior work~\cite{mannam2023framework}. Thus, we perform $150$ teleoperation experiments ($30$ tasks repeated $5$ times each) to compare against the \vone{} through \vfive{} data (including Allegro dexterous hand results as baseline) to test how \vseven{} performs on tasks outside of the tasks used during co-optimization. In Table~\ref{tab:v7_on_dash30_table}, we show that \dash{} \vseven{} performs better than Allegro and previous iterations of \dash{}, according to the number of tasks where all five out of five repetitions of each task were successful. 
The increase of 4 tasks between \vone{} and \vtwo{} is similar to that between \vfive{} and \vseven{}, showing a jump in improvement using our design and control co-optimization compared to the manual design iteration approach in prior work~\cite{mannam2023framework}. Based on overall task success (over all 150 experiments per hand), \vseven{} succeeded on $95\%$ of tasks, which is higher than the other design iterations, including \vfive{} which succeeded on $87$\% of tasks. Our simulation task of picking up and reorienting objects in-hand tested dexterity which was useful for the \dash{}-30 tasks involving holding, picking, levering, twisting, opening, and placing various objects. 

\subsubsection{Evaluation with Varying Fingertip Shape}
We also tested \vseven{}-wedge, \vseven{} with a different wedge-like fingertip shape, in simulation and teleoperation evaluation. In simulation, \vseven{} with a rounded fingertip shape performed better or the same than \vseven{}-wedge on objects \texttt{board}, \texttt{cross3d}, and \texttt{sphere} with \vseven{}-wedge scores of $0.89, 0.07, 0.42$, respectively. When tested on the real robot, we found that the wedge-like fingertip outperformed \vseven{} for \texttt{board}, \texttt{cross3d}, \texttt{sphere} objects and scored $0.89$, $0.99$, and $0.96$, respectively. This shows that fingertip shape results do not follow the trend of simulation results. 
We observed that \vseven{}'s difficulty with precision was remedied with thinner fingertips but made stability worse for \texttt{board}. 

\subsubsection{Evaluation with Non-Expert Teleoperators}
In addition to our experienced teleoperator experiments shown in Tables~\ref{tab:object_results}-\ref{tab:pose_results}, preliminary results with two non-expert teleoperators on \vthree{}, \vfive{}, and \vseven{} corroborate our findings. The two operators tested \texttt{board} on the three hands in a random order and repeated the reverse order. We then aggregated across repetitions and found that \vseven{} performed best based on maximum scores and \vfive{} performed best based on averaged scores. This is similar to Table~\ref{tab:object_results} finding that showed similar performance with \vfive{} and \vseven{} on board.

\newcommand{\dashfigheight}{0.1}

\begin{table}[t]
\centering
\setlength{\tabcolsep}{0pt}
\begin{tabular}{lccccccc}
\toprule
Robot designs & \begin{tabular}[c]{@{}c@{}}Allegro\\ \includegraphics[height=\dashfigheight\linewidth]{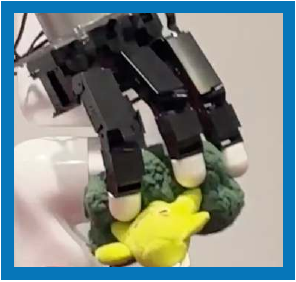}
\end{tabular} & 
\begin{tabular}[c]{@{}c@{}}v1\\ \includegraphics[height=\dashfigheight\linewidth]{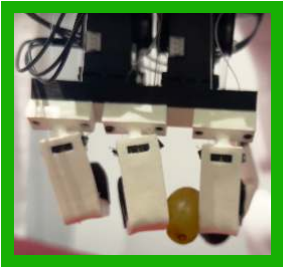}\end{tabular} &
\begin{tabular}[c]{@{}c@{}}v2\\ \includegraphics[height=\dashfigheight\linewidth]{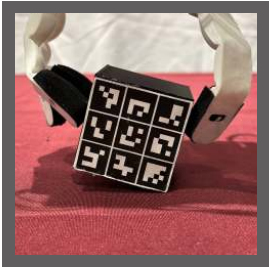}\end{tabular} &
\begin{tabular}[c]{@{}c@{}}v3\\ \includegraphics[height=\dashfigheight\linewidth]{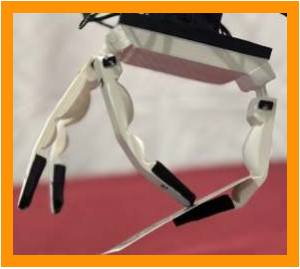}\end{tabular} & \begin{tabular}[c]{@{}c@{}}v4\\ \includegraphics[height=\dashfigheight\linewidth]{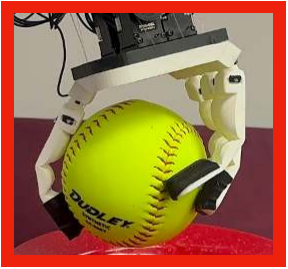}\end{tabular} & \begin{tabular}[c]{@{}c@{}}v5\\ \includegraphics[height=\dashfigheight\linewidth]{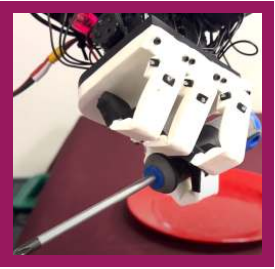}\end{tabular} & \begin{tabular}[c]{@{}c@{}}v7\\ \includegraphics[height=\dashfigheight\linewidth]{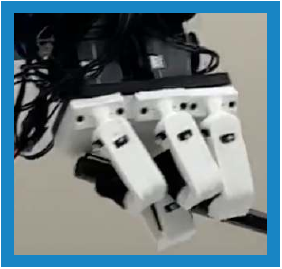}\end{tabular} \\
\midrule
\begin{tabular}[c]{@{}l@{}}Tasks with\\ 5/5 success \\ (out of 30 tasks)\end{tabular} & 7    & 10   & 14   & 16   & 17   & 19   & 23   \\ \hline
\begin{tabular}[c]{@{}l@{}}Overall success rate \\ (our of 150 tasks)\end{tabular}         & 60\% & 70\% & 82\% & 83\% & 75\% & 87\% & 95\% \\ \bottomrule
\end{tabular}
\caption{\small Task generalization of \vseven{} hand design on \dash{}-30 task suite and comparison with hand designs from prior work~\cite{mannam2023framework}.}
\label{tab:v7_on_dash30_table}
\vspace{-0.5cm}
\end{table}

%% file: Sections/Discussion.tex
\section{Discussion \& Conclusion}
\label{sec:Disc_Conc}
Using genetic algorithms and policy transfer, we were able to optimize on the design of a 16-DoF anthropomorphic soft robotic hand. 
While simulation used trained expert policies to manipulate the objects, teleoperated experiments used potentially suboptimal policies to achieve the goal pose. This is a 
limitation of our evaluation as teleoperators are not able to find the best policy for each hand which can artificially assign lower scores to a design. 
We compensated for this limitation by allowing the teleoperator to practice each task for every goal pose and object, allowing them to learn a near-optimal policy. 
Interactive grasp quality testing during teleoperation provided finer insights into the limitations and advantages of each hand design. Using only task success in both simulation and real-world experiments would have made it difficult to differentiate between hand designs. 

Despite the sim-to-real gap arising from modeling the hand as a rigid body, we saw the same trends in performance across both simulation and real world evaluations, with \vsix{} and \vseven{} scoring highest on most objects. Both optimized designs \vsix{} and \vseven{} outperformed existing designs, \vthree{} and \vfive{} from prior work~\cite{mannam2023framework}, which were a result of manual design iteration in the real-world.
While smaller scale features like finger geometry were not optimized, larger scale features like finger arrangement and palm size were successfully guided by simulation results. In contrast, features like fingertip shape did not show positive correlation across simulation and real world. A potential future direction would be to investigate the design parameters that are suitable for design optimization in simulation.
Another direction for future work involves directly deploying learned policies to evaluate hand designs in the real world to mitigate the limitations of teleoperation.

%% file: root.bbl
\begin{thebibliography}{10}
\providecommand{\url}[1]{#1}
\csname url@rmstyle\endcsname
\providecommand{\newblock}{\relax}
\providecommand{\bibinfo}[2]{#2}
\providecommand\BIBentrySTDinterwordspacing{\spaceskip=0pt\relax}
\providecommand\BIBentryALTinterwordstretchfactor{4}
\providecommand\BIBentryALTinterwordspacing{\spaceskip=\fontdimen2\font plus
\BIBentryALTinterwordstretchfactor\fontdimen3\font minus
  \fontdimen4\font\relax}
\providecommand\BIBforeignlanguage[2]{{%
\expandafter\ifx\csname l@#1\endcsname\relax
\typeout{** WARNING: IEEEtran.bst: No hyphenation pattern has been}%
\typeout{** loaded for the language `#1'. Using the pattern for}%
\typeout{** the default language instead.}%
\else
\language=\csname l@#1\endcsname
\fi
#2}}

\bibitem{piazza2019century}
C.~Piazza, G.~Grioli, M.~Catalano, and A.~Bicchi, ``A century of robotic
  hands,'' \emph{Annual Review of Control, Robotics, and Autonomous Systems},
  vol.~2, 2019.

\bibitem{feix}
T.~Feix, J.~Romero, H.-B. Schmiedmayer, A.~M. Dollar, and D.~Kragic, ``The
  grasp taxonomy of human grasp types,'' \emph{IEEE Transactions on
  human-machine systems}, 2015.

\bibitem{YCB}
B.~Calli, A.~Singh, J.~Bruce, A.~Walsman, K.~Konolige, S.~S. Srinivasa,
  P.~Abeel, and A.~M. Dollar, ``Ycb benchmarking project: Object set, data set
  and their applications,'' \emph{Journal of The Society of Instrument and
  Control Engineers}, vol.~56, no.~10, pp. 792--797, 2017.

\bibitem{Cruciani_2020}
S.~Cruciani, B.~Sundaralingam, K.~Hang, V.~Kumar, T.~Hermans, and D.~Kragic,
  ``Benchmarking in-hand manipulation,'' \emph{{IEEE} RA-L}, 2020.

\bibitem{yang2020benchmarking}
B.~Yang, P.~E. Lancaster, S.~S. Srinivasa, and J.~R. Smith, ``Benchmarking
  robot manipulation with the rubik's cube,'' \emph{IEEE RA-L}, 2020.

\bibitem{dapg}
A.~Rajeswaran, V.~Kumar, A.~Gupta, G.~Vezzani, J.~Schulman, E.~Todorov, and
  S.~Levine, ``Learning complex dexterous manipulation with deep reinforcement
  learning and demonstrations,'' \emph{RSS 2018}, 2018.

\bibitem{morgan2019learning}
A.~S. Morgan, W.~G. Bircher, B.~Calli, and A.~M. Dollar, ``Learning from
  transferable mechanics models: Generalizable online mode detection in
  underactuated dexterous manipulation,'' in \emph{ICRA 2019}, 2019.

\bibitem{cruciani_2018_dexterous}
S.~Cruciani, C.~Smith, D.~Kragic, and K.~Hang, ``Dexterous manipulation
  graphs,'' in \emph{IROS}, 2018.

\bibitem{liu2022revolver}
X.~Liu, D.~Pathak, and K.~M. Kitani, ``Revolver: Continuous evolutionary models
  for robot-to-robot policy transfer,'' \emph{arXiv preprint arXiv:2202.05244},
  2022.

\bibitem{liu2022herd}
{Xingyu Liu and Deepak Pathak and Kris M. Kitani}, ``{HERD}: Continuous
  human-to-robot evolution for learning from human demonstration,'' in
  \emph{CoRL}, 2022.

\bibitem{davidor1991genetic}
Y.~Davidor, \emph{Genetic Algorithms and Robotics: A heuristic strategy for
  optimization}.\hskip 1em plus 0.5em minus 0.4em\relax World Scientific, 1991.

\bibitem{manikas2007genetic}
T.~W. Manikas, K.~Ashenayi, and R.~L. Wainwright, ``Genetic algorithms for
  autonomous robot navigation,'' \emph{IEEE Instrumentation \& Measurement
  Magazine}, 2007.

\bibitem{vstevo2014optimization}
S.~{\v{S}}tevo, I.~Sekaj, and M.~Dekan, ``Optimization of robotic arm
  trajectory using genetic algorithm,'' \emph{IFAC Proceedings Volumes}, pp.
  1748--1753, 2014.

\bibitem{liu2023meta}
X.~Liu, D.~Pathak, and D.~Zhao, ``Meta-evolve: Continuous robot evolution for
  one-to-many policy transfer,'' in \emph{ICLR}, 2024.

\bibitem{mannam2023framework}
P.~Mannam, K.~Shaw, D.~Bauer, J.~Oh, D.~Pathak, and N.~Pollard, ``Designing
  anthropomorphic soft hands through interaction,'' in \emph{Humanoids 2023},
  2023.

\bibitem{whitman2020modular}
J.~Whitman, M.~Travers, and H.~Choset, ``Modular mobile robot design selection
  with deep reinforcement learning,'' in \emph{NeurIPS Workshop on ML for
  engineering modeling, simulation and design}, 2020.

\bibitem{chen2021codesigning}
T.~Chen, Z.~He, and M.~Ciocarlie, ``Co-designing hardware and control for robot
  hands,'' \emph{Science Robotics}, 2021.

\bibitem{chen2022c2codesign}
C.~Chen, P.~Xiang, H.~Lu, Y.~Wang, and R.~Xiong, ``$\texttt{C}^2$:co-design of
  robots via concurrent networks coupling online and offline reinforcement
  learning,'' 2022.

\bibitem{ha2017joint}
S.~Ha, S.~Coros, A.~Alspach, J.~Kim, and K.~Yamane, ``Joint optimization of
  robot design and motion parameters using the implicit function theorem.'' in
  \emph{RSS 2017}.

\bibitem{schaff2019jointly}
C.~Schaff, D.~Yunis, A.~Chakrabarti, and M.~R. Walter, ``Jointly learning to
  construct and control agents using deep reinforcement learning,'' in
  \emph{ICRA 2019}, 2019.

\bibitem{whitman2023learning}
J.~Whitman, M.~Travers, and H.~Choset, ``Learning modular robot control
  policies,'' \emph{IEEE T-RO}, 2023.

\bibitem{xu2021endtoend}
J.~Xu, T.~Chen, L.~Zlokapa, M.~Foshey, W.~Matusik, S.~Sueda, and P.~Agrawal,
  ``An end-to-end differentiable framework for contact-aware robot design,'' in
  \emph{Robotics: Science and Systems XVII}, 2021.

\bibitem{salimans2017evolution}
T.~Salimans, J.~Ho, X.~Chen, S.~Sidor, and I.~Sutskever, ``Evolution strategies
  as a scalable alternative to reinforcement learning,''
  \emph{arXiv:1703.03864}, 2017.

\bibitem{meixner2019automated}
A.~Meixner, C.~Hazard, and N.~Pollard, ``Automated design of simple and robust
  manipulators for dexterous in-hand manipulation tasks using evolutionary
  strategies,'' in \emph{Humanoids 2019}.\hskip 1em plus 0.5em minus
  0.4em\relax IEEE, 2019, pp. 281--288.

\bibitem{deimel2017automated}
R.~Deimel, P.~Irmisch, V.~Wall, and O.~Brock, ``Automated co-design of soft
  hand morphology and control strategy for grasping,'' in \emph{IROS 2017},
  2017.

\bibitem{dime_lerrel}
S.~P. Arunachalam, S.~Silwal, B.~Evans, and L.~Pinto, ``Dexterous imitation
  made easy: A learning-based framework for efficient dexterous manipulation,''
  2022.

\bibitem{videodex}
K.~Shaw, S.~Bahl, and D.~Pathak, ``Videodex: Learning dexterity from internet
  videos,'' in \emph{CoRL 2022}, 2022.

\bibitem{chen2023sequential}
Y.~Chen, C.~Wang, L.~Fei-Fei, and C.~K. Liu, ``Sequential dexterity: Chaining
  dexterous policies for long-horizon manipulation,'' 2023.

\bibitem{kim2013physically}
J.~Kim, K.~Iwamoto, J.~J. Kuffner, Y.~Ota, and N.~S. Pollard, ``Physically
  based grasp quality evaluation under pose uncertainty,'' \emph{IEEE T-RO},
  2013.

\bibitem{bonilla2014grasping}
M.~Bonilla, E.~Farnioli, C.~Piazza, M.~Catalano, G.~Grioli, M.~Garabini,
  M.~Gabiccini, and A.~Bicchi, ``Grasping with soft hands,'' in \emph{Humanoids
  2014}, 2014.

\bibitem{bauer2022towards}
D.~Bauer, C.~Bauer, A.~Lakshmipathy, R.~Shu, and N.~S. Pollard, ``Towards very
  low-cost iterative prototyping for fully printable dexterous soft robotic
  hands,'' in \emph{RoboSoft 2022}, 2022, pp. 490--497.

\bibitem{ninjaflex}
``Ninjaflex edge,''
  \href{https://ninjatek.com/shop/edge/}{https://ninjatek.com/shop/edge/},
  accessed on 2023-10-08.

\bibitem{dynamixel}
``Robotis:dynamixel-x,''
  \href{https://www.robotis.us/dynamixel-xc330-m288-t/}{https://www.robotis.us/dynamixel-xc330-m288-t/},
  accessed on 2023-10-08.

\bibitem{xarm7}
``Ufactory xarm7,''
  \href{https://www.ufactory.cc/product-page/ufactory-xarm-7}{https://www.ufactory.cc/product-page/ufactory-xarm-7},
  accessed on 2023-10-08.

\bibitem{makoviychuk2021isaac}
V.~Makoviychuk, L.~Wawrzyniak, Y.~Guo, M.~Lu, K.~Storey, M.~Macklin,
  D.~Hoeller, N.~Rudin, A.~Allshire, A.~Handa, and G.~State, ``Isaac gym: High
  performance gpu-based physics simulation for robot learning,'' 2021.

\bibitem{manus}
``Manus,'' \href{https://www.manus-meta.com}{https://www.manus-meta.com},
  accessed on 2022-11-28.

\end{thebibliography}
